\documentclass[letterpaper]{article} 
\usepackage{aaai23}  
\usepackage{times}  
\usepackage{helvet}  
\usepackage{courier}  
\usepackage[hyphens]{url}  
\usepackage{graphicx} 
\usepackage{natbib}  
\usepackage{caption} 
\usepackage{algorithm}
\usepackage{algorithmic}
\usepackage{times}
\usepackage{latexsym}
\usepackage{tabu}
\usepackage{latexsym}
\usepackage{amsmath}
\usepackage{multirow}
\usepackage{booktabs}
\usepackage[colorinlistoftodos]{todonotes}
\usepackage[inline]{enumitem}

\usepackage[hang,flushmargin]{footmisc}
\newcommand\workshopnote[1]{\renewcommand\thefootnote{}\footnote{#1}}

\usepackage{newfloat}
\usepackage{listings}
\DeclareCaptionStyle{ruled}{labelfont=normalfont,labelsep=colon,strut=off} 
\floatstyle{ruled}
\newfloat{listing}{tb}{lst}{}
\floatname{listing}{Listing}
\title{Utilizing Background Knowledge for Robust Reasoning over Traffic Situations}
\author{
Jiarui Zhang\textsuperscript{\rm 1}, Filip Ilievski\textsuperscript{\rm 1}, Aravinda Kollaa\textsuperscript{\rm 1},\\ Jonathan Francis\textsuperscript{\rm 2}, Kaixin Ma\textsuperscript{\rm 3}, Alessandro Oltramari\textsuperscript{\rm 2}
}
\affiliations{
    \textsuperscript{\rm 1}Information Sciences Institute, University of Southern California\\
    \textsuperscript{\rm 2}Human-Machine Collaboration, Bosch Research Pittsburgh\\
    \textsuperscript{\rm 3}Language Technologies Institute, Carnegie Mellon University\\

    \{jrzhang, ilievski, kollaa\} @isi.edu, \{jon.francis, alessandro.oltramari\} @us.bosch.com, kaixinm@cs.cmu.edu
}

\title{Utilizing Background Knowledge for Robust Reasoning over Traffic Situations}

\usepackage{bibentry}

\begin{document}

\maketitle

\begin{abstract}

Understanding novel situations in the traffic domain requires an intricate combination of domain-specific and causal commonsense knowledge.
Prior work has provided sufficient perception-based modalities for traffic monitoring, in this paper, we focus on a complementary research aspect of Intelligent Transportation: traffic understanding.
We scope our study to text-based methods and datasets given the abundant commonsense knowledge that can be extracted using language models from large corpus and knowledge graphs.
We adopt three knowledge-driven approaches for zero-shot QA over traffic situations, based on prior natural language inference methods, commonsense models with knowledge graph self-supervision, and dense retriever-based models. 
We constructed two text-based multiple-choice question answering sets: BDD-QA for evaluating causal reasoning in the traffic domain and HDT-QA for measuring the possession of domain knowledge akin to human driving license tests.
Among the methods, Unified-QA reaches the best performance on the BDD-QA dataset with the adaptation of multiple formats of question answers. Language models trained with inference information and commonsense knowledge are also good at predicting the cause and effect in the traffic domain but perform badly at answering human-driving QA sets. For such sets, DPR+Unified-QA performs the best due to its efficient knowledge extraction. 
\workshopnote{Accepted to Workshop on Knowledge Augmented Methods for Natural Language Processing, in conjunction with AAAI 2023.}


\end{abstract}

\section{Introduction}


\begin{figure*}[!t]
\centering
\includegraphics[width=0.6\linewidth]{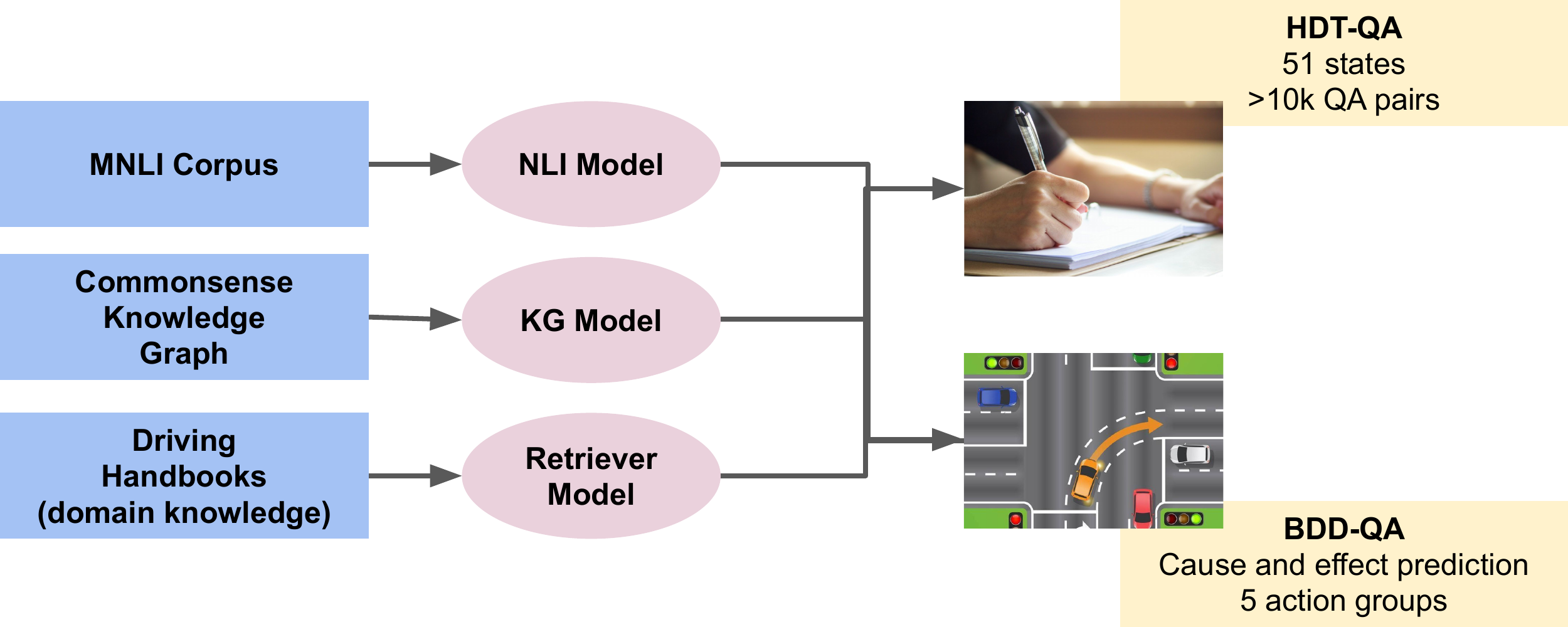}\\
\caption{Overview of our study framework, which evaluates three knowledge-enhanced language methods adapted with different knowledge sources on two evaluation datasets in a zero-shot manner.}
\label{fig:analysis}
\end{figure*}

As humans, we are able to make sense of novel situations in any domain even with limited domain-specific knowledge. The domain of intelligent traffic monitoring is a particularly attractive one, given the large market and the long list of car manufacturers that innovate in this domain~\cite{IntelligentTraffic}. Being able to understand novel situations in traffic requires a complex association of domain-specific and causal commonsense knowledge. Prior work on evaluation tasks~\cite{xu2017end, kim2018textual,xu2021sutd} and corresponding methods~\cite{halilaj2021knowledge,muppalla2017knowledge,chowdhury2021towards,wickramarachchi2020evaluation} has focused on the perception-based modality, with many of the input exemplars coming from sensors like street video cameras.

Perception (e.g., tracking of participants over time, counting cars, etc.) is certainly a key aspect of traffic monitoring. 
However, a holistic understanding of situations in traffic requires other, complementary skills such as causal and commonsense reasoning (e.g., \textit{red light causes cars to stop, direct sunlight does not}), and possession of rich domain knowledge (e.g., \textit{the driving speed in school zones is limited to 30mph}).
A large body of work has focused on general commonsense reasoning~\cite{ma2021exploring,zhang2022empirical}, but the role of commonsense and causal reasoning has not been explored thoroughly. Complementarily, ontologies for describing traffic participants~\cite{OpenXOntology} exist, but it is unclear whether connecting them to neural models can bring the desired generalizability. 
Given that natural language is a common medium for human communication, and considering the success of large language models (LMs), it is intuitive to turn to using the textual modality to develop and test robust traffic understanding models. To the best of our knowledge, understanding traffic situations presented in natural language has not been explored in depth so far. This brings up a natural question of whether large pretrained models can effectively solve realistic traffic understanding tasks dominantly presented in language, such as human driving license tests. We also observe that prior work has not systematically investigated the role of different types of background knowledge, like causal commonsense knowledge and traffic domain knowledge, in reasoning over traffic situations. 



In this paper, we focus on an evaluation for robust reasoning over traffic situations described in natural language. This allows us to reduce complexity, understand the behavior of various neural and neuro-symbolic models, and measure the impact of various sources of background knowledge. We construct two realistic text-based multiple-choice question answering (MCQA) sets: BDD-QA for evaluating causal reasoning in the traffic domain and HDT-QA for measuring the possession of domain knowledge akin to human driving license tests. 
We adopt three knowledge-driven approaches for zero-shot QA over traffic situations, based on prior methods for natural language inference, commonsense models with knowledge graph self-supervision, and dense retriever-based models. By doing so, we investigate the impact of extracting commonsense and domain knowledge from structured and unstructured data sources. An overview of our study framework is shown in Figure~\ref{fig:analysis}.




We summarize our contributions as follows:
\begin{enumerate}
    \item We formulate two novel text-based tasks and benchmarks for situational reasoning in the traffic domain: BDD-QA, which evaluates whether systems can apply causal reasoning to make decisions in arbitrary traffic situations, and HDT-QA, which tests the ability of systems to answer questions that require domain knowledge from driving manuals. We divide the datasets into meaningful partitions to allow balanced and fine-grained analysis. 
    \item We adapt different knowledge-aware methods that have been shown to generalize well across natural language reasoning tasks in prior work: natural language inference models, knowledge graph-based commonsense reasoners, and a novel retrieval-based QA method with an open-book access to official driving manuals. 
    \item We perform extensive experiments of the three methods with different language models as backbones against the two datasets in a zero-shot evaluation setup. We perform an in-depth analysis of model performance on data partitions and look closer into the model predictions to provide useful insights into the role of background knowledge for building robust traffic understanding methods.
\end{enumerate}

\section{Related work}


\subsection{Traffic Understanding}

CADP~\cite{shah2018cadp} is a spatio-temporally annotated dataset for accident forecasting using traffic camera views.
The Berkeley Deep-Drive dataset (BDD) is a
video dataset consisting of real driving videos containing abundant driving scenarios~\cite{xu2017end}. The follow-up work, BDD-X \cite{kim2018textual}, provides the action description of the BDD video and their explanations. 
TrafficQA \cite{xu2021sutd} consists of over 60K QA sets based on over 10k traffic scenes. The QA set of TrafficQA includes 6 different aspects of reasoning problems: basic understanding, attribution, introspection, counterfactual inference, event forecasting, and reverse reasoning. All QA pairs are based on visual inputs.

The traffic benchmarks have inspired corresponding methods. By using knowledge to enhance traffic understanding,~\citet{wickramarachchi2020evaluation} provides a demonstration of the process of creating and evaluating knowledge graph embedding for autonomous driving data.~\citet{chowdhury2021towards} enhance the knowledge graph for autonomous driving and the task of scene entity prediction. ITSKG~\cite{muppalla2017knowledge} is a knowledge graph framework for extracting actionable information from raw sensor data in traffic. CoSI~\cite{halilaj2021knowledge} proposes a knowledge graph-based approach for representing information sources relevant to traffic situations. 

Different from traffic monitoring, which mainly requires the perception modality, in this work, we explore a shift in the paradigm from monitoring to understanding, which requires abundant domain knowledge and commonsense reasoning methods. However, perceptual annotations provide valuable information that can be reused for creating comprehension tasks.
We utilize the textual descriptions of actions and their explanations associated with the BDD-X dataset to construct our BDD-QA dataset for causal reasoning.

\subsection{Situational Reasoning in Natural Language}

MultiNLI~\cite{williams2017broad} is a large (433k) NLI corpus with ten distinct genres of written and spoken English. MNLI has become a popular benchmark for most of the language models, such as BERT~\cite{devlin2018bert}, RoBERTa~\cite{roberta}, BART~\cite{bart}, and DeBERTa~\cite{deberta}. These methods have been widely used, but have not been applied to the traffic domain, to the best of our knowledge. In our work, we use such models that are pretrained with the MNLI dataset, to evaluate their zero-shot performance on inference tasks in the traffic domain based on the assumption that the models are capable of causal reasoning generally.

ATOMIC~\cite{sap2019atomic} provides a large atlas of everyday commonsense reasoning with a focus on inferential if-then knowledge.
CauseNet \cite{heindorf2020causenet} provides a large-scale knowledge base of claimed causal relations between concepts which can be used for casual reasoning and question-answering tasks.
\citet{Ma2021} developed a method that utilizes commonsense knowledge that is extracted from the knowledge graph to train language models via self-supervision and such models are proven to have high accuracy across several zero-shot tasks~\cite{zhang2022empirical}. Given that the knowledge-driven models perform well on many general-domain zero-shot commonsense tasks, in this work, we evaluate their performance on reasoning tasks in the traffic domain to investigate the effect of commonsense knowledge on answering traffic-domain questions.

DPR (Dense Passage Retrieval)~\cite{dpr} is a passage retrieval method using dense representations of sentences and passages for QA. Trained on a Wikipedia dump and several QA datasets, the DPR model overperforms several traditional context retrieval methods like TF-IDF and BM25.
Unified-QA~\cite{khashabi2020unifiedqa} is a generalized QA system across 4 formats of QA and is proven to perform well on all of the benchmarks. In our work, we retrieve relevant domain knowledge with DPR to help answer domain-specific questions with the Unified-QA model.

\begin{table}
\centering
\small
\caption{Examples of BDD-QA and HDT-QA. (*) denotes the correct answer. BDD-QA-EP questions ask for the correct effect given the cause and BDD-QA-CP for the correct cause given the effect. For HDT-QA, we show one example for each category (having 2,3,4,5 candidate answers)}
\label{tab: bddexaples}
\begin{tabular}{l}
\tabucline[1.1pt]\\
\textbf{BDD-QA-EP} \\
C: The light has turned green and traffic is flowing smoothly. \\
E1: The driver jerks abruptly to the right; \\
E2: The car is driving straight into the parking lot; \\
E3: The car accelerates slowly to a maintained speed(*)\\\tabucline[1.1pt]\\
\textbf{BDD-QA-CP} \\
E: The car is driving forward slowly. \\
C1: The light ahead was red at the intersection; \\
C2: The passenger is seated in the car and the lane is clear; \\
C3: There is slow traffic in front of the car(*)\\\tabucline[1.1pt]\\
\textbf{HDT-QA~(2 candidates)} \\
Q: Yield also means stop if you cannot merge safely\\ into the flow of traffic\\
A1: True(*);
A2: False\\\tabucline[1.1pt]\\
\textbf{HDT-QA~(3 candidates)} \\
Q: You just sold your vehicle. \\You must notify the DMV within \underline{~~~~~} days.\\
A1: 15;
A2: 5(*);
A3: 10;\\\tabucline[1.1pt]\\
\textbf{HDT-QA~(4 candidates)} \\
Q: You have the right of way when you are:\\
A1: Entering a traffic circle;\\
A2: Backing out of a driveway;\\
A3: Leaving a parking space;\\
A4: Already in a traffic circle(*)\\\tabucline[1.1pt]\\
\textbf{HDT-QA~(5 candidates)} \\
Q: Why should you cover your cargo during its transportation?\\
A1: To protect the cargo from weather only;\\
A2: There is no reason;\\
A3: To protect people from the cargo's contents if it spills;\\
A4: To protect people from the cargo's contents if it spills\\ and to protect the cargo from the weather(*);\\
A5: To help your vehicle accelerate smoothly\\ \tabucline[1.1pt]\\
\end{tabular}
\label{tab:examples}
\end{table}

\section{Tasks \& Datasets}

We refer to situational reasoning as the task of providing the best possible justification or decision for a given situation. For instance, in the traffic domain, the justification for ``Why did the car accelerate?'' could be the light turning green or the absence of traffic. We focus on situational reasoning in the traffic domain. Formally, we pose traffic situational reasoning as a multiple-choice question-answering task. Each task entry consists of a natural language question $Q$, and $n$ candidate answers $\{A_{1}, ..., A_{n}\}$, consisting of $1$ correct answer and $n-1$ distractors. The task of an AI model is to select the most probable answer among the candidates.

To evaluate whether AI models comprehensively understand the traffic domain knowledge, we introduce two diverse tasks: BDD-QA and HDT-QA. BDD-QA tests model understanding of the causality of traffic participant behaviors and their reasons, while HDT-QA focuses on driving policies, and legible or recommended drivers' decisions. Examples of the two datasets are shown in Table~\ref{tab: bddexaples}. Next, we describe the details of these two QA datasets.

\subsection{BDD-QA dataset}
\label{sec:bddset}

\textbf{Data Source}~~We create a multiple choice question set from the Berkeley Deep-Drive Explanation Dataset \cite{kim2018textual} (BDD-X), which includes over 26k QA sets and covers an extensive range of weather and road types. Given that the dataset consists of an action and its justification (reason), we treat the action as an effect and the justification as a cause.

\textbf{Sampling Strategy} We use sentence transformers~\cite{reimers2019sentencebert} to represent the sentences as sentence embeddings. The model we use, \textit{all-mpnet-base-v2}, is based on MPNet~\cite{song2020mpnet} and is pretrained on over 1B sentence pairs and reportedly reached a state-of-art performance. All of the sentence embeddings are normalized and we compute their distance using cosine similarity. To make the sentence embedding more accurate, we transform each sentence into a declarative and grammatically complete one. For example, if a justification sentence is \textit{to turn left}, it will be transformed to \textit{The car wants to turn left} by filling in the subject. If the justification sentence is \textit{because the light turns red}, we will remove the \textit{because}, resulting in the sentence \textit{The light turns red}. To avoid having causes that are too similar, we remove the QA pairs that have a cause that has a similarity to others higher than a threshold, in our work, we set this threshold to \textit{0.9}.

\begin{table}
\centering
\small
\caption{Examples of 5 classes of effects (car actions) discovered in the BDD-QA dataset.}
\label{tab:classexaples}
\begin{tabular}{l r}

\toprule[1.1pt]
\bf Classes& \bf Sentences\\\toprule[1.1pt]
\multirow{3}*{\textbf{Accelerate}} &The car accelerates\\
&The car inches forward\\
&The car is traveling down the road. \\\midrule[1.1pt]
\multirow{3}*{\textbf{Slow}} 
&The car slows slightly \\
&The car moves forward slowly \\
&The car maintains a slow speed \\\midrule[1.1pt]
\multirow{3}*{\textbf{Stop}} 
&The car stops \\
&The car is parked at the right curb  \\
&The car is stationary\\\midrule[1.1pt]
\multirow{3}*{\textbf{Merge}} 
&The car merges into the lane to its left \\
&The car is moving to the left lane \\
&The car is switching lanes to the left \\\midrule[1.1pt]
\multirow{3}*{\textbf{Turn}} 
&The car slows down and pulls to the right \\
&The car turns left and drives forward \\
&The car swerves to the left and slows \\\midrule[1.1pt]
\end{tabular}
\label{tab:examples}
\end{table}

To generate multiple-choice questions, for each cause-and-effect pair, we employ two different approaches: (1) Sampling QA-pairs by dissimilar effects for effects-prediction, \textbf{(BDD-QA-EP)}, and (2) Sampling QA-pairs by dissimilar causes for causes-prediction, \textbf{(BDD-QA-CP)}.
We randomly sample cause/effect from other pairs, then we set an upper bound of sentence similarity to avoid sampling too similar answers that are also correct for a question. The upper bound threshold $t$ is set to 0.4 in our work, all the randomly sampled cause/effect whose similarity with the true answer is higher than the threshold would be removed and we will re-sample until it satisfies the threshold restriction.
For example, given the cause \textit{The road doesn't have much traffic} and effect \textit{The car accelerates down the road}, a good effect distractor would be \textit{The vehicle is stopped} rather than \textit{The car is accelerating faster}. To improve the diversity of the sampled distractors, we also make sure that the dissimilarity between the distractors is also lower than the same threshold $t$. 

\textbf{Action Classification}~~The project ~\cite{OpenXOntology} presents an ontology that describes the concepts in the traffic domain, including car maneuvers. In the real world, the possible actions performed by cars are few and lead to a limited set of effects. Inspired by this, we classify the effects of BDD-QA into several classes, each of which presents a type of car action. After we compute the sentence embeddings for each action, we use k-means~\cite{macqueen1967classification} for the action sentence clustering. After trying different numbers of classes, we noticed that the clusters are intuitive. We decide to use 5 classes that can represent the car's actions best from a human perspective, namely: \textit{accelerate}, \textit{slow}, \textit{stop}, \textit{merge} and \textit{turn}. 
Table~\ref{tab:classexaples} shows examples for each of the five classes.

\subsection{HDT-QA dataset}

We also develop a complementary QA set by using driving tests intended to test the driving knowledge of humans. We call this dataset HDT-QA (Human Driving Test QA).

\textbf{Data Source}~~We scrap the questions of HDT-QA from a website that contains the driving test data and driving manuals from all 51 states of the USA~\cite{Drivingtest}. For each state, there are three subjects of driving tests and manuals \textit{Motorcycle}, \textit{Car} and \textit{Commercial
Driver}, resulting in 153 state-subject combinations. The test data are all multiple-choice questions, where the number of answer candidates ranges between 2 and 5. The statistics of our HDT-QA set for each number of candidate answers are shown in Table~\ref{tab: statistic}.

\textbf{Filtering}~~To make our test data fair for text-based methods, we only keep questions that can fully be answered based on the textual content. We exclude all questions with images (e.g., traffic signs). We have also filtered out the questions where one of the candidate answers is \textit{all of the above}, as answering such questions is not the focus of our work.


\begin{table}[!h]
	\centering
	\small
	\caption{Statistics of the partitions of our two datasets.}
	\label{tab: statistic}
	\begin{tabular}{l| l r }
	\hline
        \bf Dataset&\bf Partition&\bf Size\\\hline
        \multirow{2}*{\bf{BDD-QA}}&Event Prediction&3,139\\
        &Cause Prediction &3,139\\\hline
        \multirow{4}*{\bf{HDT-QA}}&2 answers&131\\
        &3 answers&2,398\\
        &4 answers&7,558\\
        &5 answers&40\\\hline
	\end{tabular}
\end{table}
\section{Method}
\label{sec:methods}

A key strength of neural language models is their ability to provide answers for open-world reasoning tasks~\cite{devlin2018bert,radford2019language}. While the models can be tuned on a dataset to enhance their accuracy, prior work has shown that this leads to over-fitting~\cite{ma2021exploring}. Anticipating similar over-fitting to QA sets in the traffic domain, we focus on developing models that include commonsense and domain knowledge to enable them to better understand situations in traffic and generalize across unseen traffic QA tasks.  
We propose three methods to enrich language models with background knowledge for the traffic domain: \textit{NLI-based models}, \textit{Knowledge-based models}, and \textit{Retrieval-based models}. We use readily available models and run them directly on our two benchmarks in a zero-shot setting.

\subsection{Inference-based reasoning models}

We propose an NLI method based on semantic-level reasoning for the zero-shot evaluation of our tasks. Given a premise and its hypothesis that can be from any domain, the NLI model we use is expected to capture coherence information between them. It takes two sentences as the input and outputs a 3-digit vector which represents confidence letting the sentence pair's relation be entailment, neutral, and contradiction.

To avoid the effect of the data format, we keep our QA data as declarative sentence pairs. As is described in the last paragraph, given a sentence pair $S_1, S_2$, the model is asked to give the score of entailment, neutral, and contradiction of these two sentences as $P_e, P_n, P_c$, respectively. 
Then the model chooses the candidate with the highest margin score of $P_e-P_c$ as the true answer.
Formally, the final prediction of the model will be:
$$
pred=
\arg \max_{i}(P(e|p_{i},MNLI)-P(c|p_{i},MNLI))
$$
Where $p_i$ is the $i_{th}$ candidate sentence pair for the model to choose. $e$ stands for entailment and $c$ stands for contradiction. $MNLI$ refers to the inference knowledge that is extracted from the MNLI dataset by the language model.

\subsection{KG-based reasoning models}
\label{sec: kgmethod}

Self-supervision with structured knowledge has proven to improve the language model's ability to solve reasoning problems in a zero-shot setting~\cite{Ma2021,zhang2022empirical,dou2022zero}. Thus, we evaluate the performance of language models enhanced by commonsense knowledge of traffic domain question-answering problems. 

Our knowledge-based QA framework follows the setup of~\citet{ma2021exploring,zhang2022empirical}. Given a question and several candidates, the model chooses the question-candidate pair that best entails the knowledge extracted from the knowledge graph:

$$
pred=\arg \max_{i}(R(s_{i}|CSKG))
$$
where $R(s|CSKG)$ is the reasonable score of a sentence based on the knowledge that the language model extracted from the Commonsense Knowledge graph.

We chose models that are pretrained on a synthetic dataset that is generated from the Commonsense Knowledge Graph~\cite{ilievski2021cskg}. In total, Commonsense Knowledge Graph contains 7 million commonsense statements, which is a significant portion of implicit facts that the commonsense models have absorbed via self-supervision.

\subsection{Retrieval-based QA models}

Retrieval-based QA models aim to find evidence for the correct answer for a given contextualized question in background knowledge documents.
This task aims to capture domain knowledge such as rules and regulations we abide by while driving in traffic. We pose the input to the retrieval-based models as a multiple-choice question answering based on context. The main idea is to extract the most relevant content as evidence that can help answer the question based on domain documents and use it as a context to find the best possible choice.

We propose a novel pipeline for domain-based question answering. We use Dense Passage Retriever (DPR) \cite{dpr} as the relevant domain extractor module to encode each paragraph and question sentence into vectors. Given a corpus $C$ include several passages $PA$, we use a cosine similarity score to extract the most relevant paragraph:
$$
PA_{select} = \arg \max_{PA_i}~cos\_sim(PA_i,Q), PA_i \in C
$$

The passage extracted then along with the question and choices are passed to a QA model, then the QA model generates the predicted answer, the probability of generating an answer is:
$$
\begin{aligned}
    &P(GA|Q,A,C) = \\
    &\prod_{i=i}^n P(ga_i|Q,A,PA_{select},ga_1,ga_2,...,ga_{i-1})
\end{aligned}
$$
Where $C$ is the whole corpus, for our task, we use a comprehensive collection of driving manuals as the domain corpus~\cite{Drivingtest}. After processing the manual PDFs, we divide them into sentences by simply using the punctuation \textit{" . ? ! "} as sentence boundaries. Then we use a T5-based grammar correction \cite{GrammarCorrection} as a data cleaning tool to make our corpus have correct spelling and grammar. Finally, we combine every 10 neighborhood sentences as paragraphs that are fed to the QA model.\footnote{Paragragh segmentation based on the similarity between neighboring sentences yielded consistently worse results.} In total, there are $42.979$ paragraphs and each paragraph has an average word count of $147.5$.

After the model generates the predicted answer, we choose the answer which is most similar to the $GA$ as our final prediction.

\begin{table*}[!ht]
	\centering
	\small
	\caption{Evaluation results on the BDD-QA-EP and BDD-QA-CP dataset of our three methods. The Vanilla Roberta-large model was only pretrained by unlabelled corpus. The Vanilla Unified-QA method doesn't feed the most relevant corpus to the model. Different from other results, supervised models only use 10\% of the whole data for evaluation. The result of the human evaluation is included in the last row.}
	\label{tab: main result}
	\begin{tabular}{ll | rrr }
	\hline
	    {\bf Method} & \bf Model & BDD-QA-EP& BDD-QA-CP&~~~~~~~~~~~~Avg\\\hline
        \bf Random & - &33.3&33.3&33.3\\ \hline
        \multirow{2}*{\bf Vanilla models} & Roberta-large &36.1&43.6&40.0 \\ 
        &{UnifiedQA-v2}&\bf{75.9}&63.3&\bf{69.6}\\
        \hline
        \multirow{5}*{\bf NLI-based}
        &Roberta-large-mnli&63.2&58.6&60.9\\
        &Bart-large-mnli&66.3&56.8&61.5\\
        &Deberta-large-mnli&67.0&54.5&60.8\\
        &Deberta-v2-xlarge-mnli&71.3&59.7&65.5\\
        &Deberta-v2-xxlarge-mnli&72.4&\bf{64.9}&68.7\\\hline
        \multirow{5}*{\bf Knowledge-based}
        &Roberta-base&63.3&54.4&58.9\\
        &Roberta-large&71.7&56.8&64.3\\
        &T5-small&54.3&46.0&50.1\\
        &T5-large&66.4&55.5&61.0\\
        &T5-3b&71.0&58.6&64.8\\\hline
        \bf Retrieval-based& {UnifiedQA-v2 + DPR}&71.4&55.7&63.6\\\hline
        {\bf Supervised}&Roberta-large&93.0&81.2&86.1\\\hline
        {\bf Human}&-&74.0&68.0&71.0\\\hline
	\end{tabular}
\end{table*}

\section{Experiments}

\subsection{Model setup}

For Inference-base reasoning models, we select several NLI models that are trained on the MNLI~\cite{williams2017broad} dataset. These models are: \textit{Roberta-large}, \textit{Bart-large}, \textit{Deberta-large}, \textit{Deberta-v2-xlarge} and \textit{Deberta-v2-xxlarge}. 

For KG-based reasoning models, we choose the five models from \citet{zhang2022empirical} for evaluation: Roberta-base, Roberta-large and T5-small, T5-large, and T5-3b models. 

For Retrieval-based models, we use the sentence transformer model pretrained in \cite{dpr} to encode the paragraphs and sentences. As a QA model, we use the Unified-QA T5-based transformer model \cite{uqa}, to output the most probable candidate answer. 

\subsection{Other Baselines}

To indicate the upper bound of models' performance on our dataset, we have also tested a supervised method based on tuning Roberta-large. We have divided all of the datasets into training and testing parts with a ratio of 90\% and 10\%. For the HDT-QA dataset with 2 and 5 questions, we do not perform the supervised method due to the limited size of the dataset.

To investigate the correlation between our two datasets, we also include a transfer learning experiment that takes the supervised model trained on one dataset to evaluate the other dataset. We also report the results of an unsupervised vanilla Roberta-large model and vanilla Unified-QA model as well as of a random baseline.

\begin{table*}[!ht]
	\centering
	\small
	\caption{Evaluation results on the BDD-QA-EP dataset of our models on 5 action classes.}
	\label{tab: class result}
	\begin{tabular}{cc | ccccc}
	\hline
	    \multicolumn{2}{c|}{\multirow{2}{*}{\bf Methods}}& \multicolumn{5}{c}{\bf Action Class}\\
        && Accelerate& ~~Slow~~~& ~~Stop~~~& ~~Merge~~& ~~Turn~~~\\\hline
        \multirow{5}*{\bf NLI-based}
        &Roberta-large&70.5&59.7&68.5&59.6&51.8\\
        &Bart-large&77.1&59.5&67.5&68.6&57.1\\
        &Deberta-large&74.7&58.4&71.6&70.1&58.7\\
        &Deberta-v2-xlarge&74.5&62.5&\textbf{76.2}&74.5&66.2\\
        &Deberta-v2-xxlarge&\textbf{79.0}&65.3&75.8&72.2&65.7\\
        \hline
        \multirow{5}*{\bf Knowledge-based}
        &Roberta-base&56.5&57.4&51.5&66.5&\textbf{82.4}\\
        &Roberta-large&68.7&\textbf{70.4}&68.5&\textbf{77.0}&81.3\\
        &T5-small&57.1&58.8&59.0&42.6&58.0\\
        &T5-large&59.7&60.8&65.8&65.5&80.1\\
        &T5-3B&65.1&68.7&63.7&75.1&81.8\\\hline
	\end{tabular}
\end{table*}

\section{Results}

We have tested three methods on our two datasets. We use accuracy as our evaluation metric. For the BDD-QA dataset, the Unified-QA model reaches the best performance by adapting multiple formats of question answers. For the HDT-QA dataset, our proposed retrieval-based model, DPR+Unified-QA outperforms other methods by nearly 20 points, which shows the effectiveness of retrieving domain knowledge for question answering.

\subsection{BDD-QA Task}

Table~\ref{tab: main result} shows the results for three different types of models (NLI-based, KG-based, and Retrieval-based QA) on the BDD-QA dataset. The result shows that the best accuracy is obtained by Unified-QA-v2, which benefits from its adaption of multiple formats of question answers. Such performance is 29.6 points higher than the unsupervised Roberta-large result, but still, has a large gap with the supervised model (16.5 points). In general, NLI-based models have a balanced performance in the two datasets while KG-based models perform well on the BDD-QA-EP dataset but poorly on BDD-QA-CP.

Zero-shot performance depends on model size and background training data.
We see that the performance of NLI-based models is getting continuously better when the model size grows, which is expected. Specifically, NLI-based models' performance on BDD-QA-EP is growing faster when the model size is smaller, while for BDD-QA-CP the growth is more obvious between bigger models. Given the sentence embeddings of all the causes and effects, we have computed the average sentence similarity between causes as $0.403$, while for effects it is $0.553$, which means causes are more different from each other than the effects. The difference of similarities is intuitive because the effects (cars' actions) are limited while the causes represent factors in the environment and are therefore more open-ended. By focusing on three Deberta-large models from Table~\ref{tab: main result}, we observe that the performance on BDD-QA-CP grows faster than BDD-QA-EP when the model size is bigger, $(64.9-59.7>72.4-71.3$ while $59.7-54.5 \approx 71.3-67.0)$.
 Such a result implies that predicting the correct cause given an effect needs a language model having a larger capacity, which is coherent with the observation between cause and effect prediction.

Among all of the knowledge-based models, T5-3b, the best-performing model across five zero-shot benchmarks in~\citet{zhang2022empirical}, which is an encoder-decoder model trained with sequence-to-sequence language model loss, achieves the best performance. Roberta-large, the encoder-only model trained on masked language model loss, also reaches a high performance that is close to T5-3b's. Overall, knowledge-based models are not good at predicting the cause, none of them reaches 60\% accuracy in the BDD-QA-CP dataset. To explain this low performance, we look deeper into the properties of the synthetic training data that was constructed in~\cite{Ma2021} from CSKG~\cite{ilievski2021cskg}. The source graph has abundant information about X causes Y, which we translate into training questions of the form: \textit{What does X cause?} then the different effects are sampled. However, there are no questions about the inverse formulation \textit{What is the cause of X?} like what we do to the BDD-QA-CP dataset. Such a result is consistent with the finding in~\citet{zhang2022empirical} that the language models learn to answer well about what is in the data and do not generalize well to other aspects. Finally, there is still a large gap between both models’ accuracy and that of the supervised Roberta model, which leaves plenty of space for future improvement of the models.

Regarding the retriever-based models, Unified QA itself benefits from its adaption of multiple formats of question-answer pairs so its performance is close to the best one among the NLI-based and KG-based models. However, after we provide the most relevant passage with it, the performance decreases. This implies that knowledge in the driving handbook provides little help for the model to answer the BDD-QA questions.

\begin{table*}[!ht]
	\centering
	\small
	\caption{Evaluation results on the HDT-QA dataset of 10 models. We only train the HDT-QA-3Q and HDT-QA-4Q for supervised learning because the size of HDT-QA-2Q and HDT-QA-5Q are both limited.}
	\label{tab:hdtresult}
	\begin{tabular}{cc | cc cc c}
	\hline
	    \multicolumn{2}{c|}{\bf Methods} & HDT-QA-2Q & HDT-QA-3Q &HDT-QA-4Q & HDT-QA-5Q&~~Avg~~\\\hline
        \bf Random & - &50.0&33.3&25.0&20.0&25.7\\ \hline
        \multirow{2}*{\bf Vanilla models}
        &Roberta-large&55.4&43.7&36.7&9.10&36.2\\
        &UnifiedQA-v2&67.9&51.6&44.0&25.0&47.1\\\hline
        \multirow{5}*{\bf NLI-based}
        &Roberta-large-mnli&63.4&41.0&33.0&15.0&38.1\\
        &Bart-large-mnli&65.6&46.1&34.3&22.5&42.2\\
        &Deberta-large-mnli&62.6&42.8&34.5&15.0&38.7\\
        &Deberta-v2-xlarge-mnli&62.6&45.7&36.3&17.5&40.5\\
        &Deberta-v2-xxlarge-mnli&63.4&47.5&36.1&22.5&42.4\\\hline
        \multirow{5}*{\bf Knowledge-based}
        &Roberta-base&60.3&36.0&29.0&25.0&37.6\\
        &Roberta-large&63.4&38.9&30.8&27.5&40.1\\
        &T5-small&57.3&35.4&23.8&22.5&34.7\\
        &T5-large&64.1&40.3&24.6&22.5&37.9\\
        &T5-3b&66.4&45.0&25.1&22.5&39.7\\\hline
        \bf Retrieval-based& {UnifiedQA-v2 + DPR}&\textbf{80.2}&\textbf{57.3}&\textbf{47.7}&\textbf{60.0}&\textbf{61.3}\\\hline
        \bf Supervised & Roberta-large & - &71.7& 75.8 &-&73.8 \\ \hline
        {\bf Transfer learning }&Roberta-large&45.8&43.2&37.9&21.8&37.2\\\hline
	\end{tabular}
\end{table*}

\textbf{Overlap analysis.} We expect that the language model performance highly depends on their training data and training methods. To investigate the predictions of two different kinds of models, we look closer at the predictions of the best-performing NLI-based model and KG-based model Deberta-v2-xxlarge and Roberta-large, respectively. Among 3,139 QA pairs in BDD-QA-EP, Deberta-v2-xxlarge can correctly answer 2,256 questions and Roberta can predict 2,302 QA pairs correctly. However, there are only 1,826 answers that are correctly answered jointly, which means that for an ensemble model that can combine the two models' correct answers, the accuracy would be 87.0\% on BDD-QA-EP. For BDD-QA-CP, the combined accuracy would be 79.1\%. Such results are much higher than the performance of the two models alone and are close to the result of supervised learning. Such results encourage us to investigate the granular performance of two kinds of models by splitting the dataset into several classes. We discuss this next.

\textbf{Result on classes of effects.} 
We investigate the granular performance of different models on the action (cause) classes of the BDD-QA-EP dataset. Table~\ref{tab: class result} shows that the Knowledge-based models perform best on the \textit{Turn}, \textit{Merge} and \textit{Slow} classes, while NLI-based models do better on the other two classes (\textit{Move} and \textit{Stop}). In the real world, turning, merging, and slowing are more complicated actions than moving and stopping because their causes vary a lot. 
For example, if a car merges to the left, the cause might be \textit{The snow is taking up too much of the right lane}, or \textit{There are pedestrians in the right} or \textit{The car is preparing to execute a u-turn maneuver}, or a long list of other causes. But for Accelerate, the list of causes is more narrow: it would either be \textit{The light turned green} or \textit{The road is clear}.
We conclude that knowledge-based models can handle complex reasoning tasks in the traffic domain, while NLI-based models are better at surface-level inference. Here, again we see that the performance for most action classes grows with the increase in model size. 

\textbf{Human evaluation. }
Since BDD-QA is constructed by an automatic procedure, we perform a human evaluation for this dataset. For each partition of the dataset, we randomly select 50 QA pairs. Besides providing an answer for each question, we also ask the annotators to give confidence in their results on a 5-point Likert scale (1 being low and 5 being high) to check whether the questions are commonsensical. Each question is answered by 3 humans and we choose the majority-voted candidate as the correct answer. The result is shown in the last row of Table~\ref{tab: main result}. We note that the human score is 1.4\% points higher than the best zero-shot result, and the average annotator confidence is $3.67$. We are surprised by the relatively low human score on this task. Upon further analysis, we notice that most of the questions that are wrongly answered by humans have multiple reasonable answers, e.g., given the cause \textit{The cross traffic has cleared and the pedestrian is gone} The correct effect is \textit{The car makes the left turn and heads down the street}, but another effect \textit{The car very slow accelerates} is also reasonable given the cause. For this question, all three humans chose the latter answer. In the future, we will revise the dataset to further clean up the noise.


\subsection{HDT-QA Task}

As shown in Table~\ref{tab:hdtresult}, all of the NLI-based and KG-based models perform consistently poorly on the HDT-QA dataset. The growth of the performance together with the model size that was observed on the BDD-QA dataset is not as clearly visible on the HDT-QA dataset. Such a result implies that obtaining basic commonsense knowledge and semantic inference information is not enough for answering human driving test questions. The examples of HDT-QA, shown in Table~\ref{tab:classexaples}, are about the details of the policies of driving. It is hard even for humans to pass the driving exams without learning specific rules and driving policies captured in domain manuals. As a result, it is intuitive to retrieve domain knowledge to help language models better answer the human driving test.

As shown in Table~\ref{tab:hdtresult}, UnifiedQA itself performs slightly better than the NLI-based and KG-based models. After we provide the most relevant paragraph to the QA model automatically with the DPR retriever, the average performance of the subsets improves by 14.2\% points. This means the QA system is able to capture relevant domain knowledge efficiently for answering human driving test questions, especially when combined with a dense retrieval model that can provide the most relevant set of sentences that serve as evidence.

The performance of unsupervised learning and transfer is close to that of NLI-based and KG-based models, which means the NLI, KG, and BDD-enhanced models contribute little for models to answer the HDT-QA questions. Lastly, supervised learning reaches a high accuracy, which is intuitive because some of the questions in the training and the test set share the same knowledge.
\section{Discussion}

Our experiments show that language models trained with semantic level coherence information and commonsense knowledge extracted from knowledge graphs are both performing well at answering traffic domain reasoning tasks. Careful domain information extraction and model adaption leads to significant improvement of language models' performance on domain-specific QA sets. Next, we discuss and list future extensions inspired by our results.

\textbf{More efficient retrieving of domain knowledge}
In this paper, we use Dense Passage Retrieval to capture the most relevant paragraph in the driving manuals and directly fed the paragraph to the QA system. However, the language models are not learning the policies, they are answering the questions only by consulting the driving manual simultaneously in an open-book setting, which can be misled by vague information or context mismatch. Developing methods for effectively extracting domain knowledge and letting language models learn this knowledge is an important future work for solving HDT-QA and similar traffic-specific QA tasks. Selecting informative sentences or building knowledge graphs based on raw data are possible methods for a more precise injection of domain knowledge.

\textbf{Joint reasoning over domain knowledge and causal commonsense knowledge} Our methods have isolated the different knowledge types and have focused on learning either causal or domain knowledge, but not both. We anticipate that a robust AI model should be able to reason over both knowledge types simultaneously and know which questions require which specific knowledge statements. A key future item is integrating the two knowledge types together in a joint reasoning system and developing benchmarks that can evaluate both aspects in a more coherent manner.

\textbf{From natural language QA to multi-modal traffic QA}
Our study focused on understanding traffic situations based on using only text as background information.
However, visual information in images and videos is a key component of traffic scenes. Automatic driving systems take images as input and make decisions according to them. Popular datasets like Traffic-QA~\cite{xu2021sutd} are also based on background information that is given by an image.
Considering visual information is necessary for building a holistic robust system that understands traffic situations. Thus, we believe that multi-modal benchmarks and methods that combine perception and background knowledge for situational traffic understanding are the next milestones toward developing real-world neuro-symbolic traffic models.

\section{Conclusion}


Besides robust perception models, a holistic situational understanding of traffic requires generalizable models equipped with commonsense and domain knowledge, which has received little attention in prior work. To bridge this gap, this paper studied the strengths and weaknesses of current zero-shot adaptation methods and their granular effects on different partitions of tasks.
We constructed two benchmarks for evaluating the ability of zero-shot causal reasoning (BDD-QA) and accessing domain knowledge (HDT-QA) to answer questions about traffic situations. 
We proposed two representative methods that focus on zero-shot causal reasoning and one novel pipeline for retrieving domain information.
On the BDD-QA dataset, we have observed the limited prediction overlap between the two models. To deeper investigate that, we split
the actions of the cars into several categories, observing significant differences in performance between the methods, which provides a convincing explanation of the result of the limited overlap.
On the HDT-QA dataset, we observed a giant performance improvement after we fed the relevant information extracted from driving manuals in an open-book setting. 

The methods in our work can be easily extended to other traffic and related tasks in other domains. Our code and data can be downloaded at \url{https://github.com/saccharomycetes/text-based-traffic}.  Key future directions include more efficient knowledge retrieval, better integration of causal and domain knowledge, and the development of multi-modal traffic benchmarks and methods.


\section{Acknowledgements}
We would like to thank all of the anonymous reviewers for their valuable suggestions. We also thank Zhivar Sourati, Yifan Jiang, Jiachen Zhang, Thiloshon Nagarajah, Xiaoshu Luo, Riley Li, and Prateek Chhikara for participating in our human evaluation study.


\bibliography{aaai23}

\begin{thebibliography}{31}
\providecommand{\natexlab}[1]{#1}

\bibitem[{Ope(2020)}]{OpenXOntology}
 2020.
\newblock ASAM OpenXOntology.
\newblock \url{https://www.asam.net/project-detail/asam-openxontology/}.

\bibitem[{Gra(2021)}]{GrammarCorrection}
 2021.
\newblock Fine-Tune a Transformer Model for Grammar Correction.
\newblock \url{https://www.vennify.ai/fine-tune-grammar-correction/}.

\bibitem[{Dri(2022)}]{Drivingtest}
 2022.
\newblock Free DMV Permit Practice Tests.
\newblock \url{https://www.dmv-test-pro.com/}.

\bibitem[{Int(2022)}]{IntelligentTraffic}
 2022.
\newblock Intelligent Traffic Management System Market Worth \$27.92 Billion By
  2030.
\newblock
  \url{https://www.grandviewresearch.com/press-release/global-intelligent-traffic-management-system-market}.

\bibitem[{Chowdhury et~al.(2021)Chowdhury, Wickramarachchi, Gad-Elrab,
  Stepanova, and Henson}]{chowdhury2021towards}
Chowdhury, S.~N.; Wickramarachchi, R.; Gad-Elrab, M.~H.; Stepanova, D.; and
  Henson, C.~A. 2021.
\newblock Towards Leveraging Commonsense Knowledge for Autonomous Driving.
\newblock In \emph{ISWC (Posters/Demos/Industry)}.

\bibitem[{Devlin et~al.(2018)Devlin, Chang, Lee, and
  Toutanova}]{devlin2018bert}
Devlin, J.; Chang, M.-W.; Lee, K.; and Toutanova, K. 2018.
\newblock Bert: Pre-training of deep bidirectional transformers for language
  understanding.
\newblock \emph{arXiv preprint arXiv:1810.04805}.

\bibitem[{Dou and Peng(2022)}]{dou2022zero}
Dou, Z.-Y.; and Peng, N. 2022.
\newblock Zero-shot Commonsense Question Answering with Cloze Translation and
  Consistency Optimization.
\newblock \emph{arXiv preprint arXiv:2201.00136}.

\bibitem[{Halilaj et~al.(2021)Halilaj, Dindorkar, L{\"u}ttin, and
  Rothermel}]{halilaj2021knowledge}
Halilaj, L.; Dindorkar, I.; L{\"u}ttin, J.; and Rothermel, S. 2021.
\newblock A knowledge graph-based approach for situation comprehension in
  driving scenarios.
\newblock In \emph{European Semantic Web Conference}, 699--716. Springer.

\bibitem[{He et~al.(2020)He, Liu, Gao, and Chen}]{deberta}
He, P.; Liu, X.; Gao, J.; and Chen, W. 2020.
\newblock DeBERTa: Decoding-enhanced BERT with Disentangled Attention.

\bibitem[{Heindorf et~al.(2020)Heindorf, Scholten, Wachsmuth, Ngonga~Ngomo, and
  Potthast}]{heindorf2020causenet}
Heindorf, S.; Scholten, Y.; Wachsmuth, H.; Ngonga~Ngomo, A.-C.; and Potthast,
  M. 2020.
\newblock Causenet: Towards a causality graph extracted from the web.
\newblock In \emph{Proceedings of the 29th ACM International Conference on
  Information \& Knowledge Management}, 3023--3030.

\bibitem[{Ilievski, Szekely, and Zhang(2021)}]{ilievski2021cskg}
Ilievski, F.; Szekely, P.; and Zhang, B. 2021.
\newblock Cskg: The commonsense knowledge graph.
\newblock In \emph{European Semantic Web Conference}, 680--696. Springer.

\bibitem[{Karpukhin et~al.(2020)Karpukhin, Oğuz, Min, Lewis, Wu, Edunov, Chen,
  and Yih}]{dpr}
Karpukhin, V.; Oğuz, B.; Min, S.; Lewis, P.; Wu, L.; Edunov, S.; Chen, D.; and
  Yih, W.-t. 2020.
\newblock Dense Passage Retrieval for Open-Domain Question Answering.

\bibitem[{Khashabi, Kordi, and Hajishirzi(2022)}]{uqa}
Khashabi, D.; Kordi, Y.; and Hajishirzi, H. 2022.
\newblock UnifiedQA-v2: Stronger Generalization via Broader Cross-Format
  Training.

\bibitem[{Khashabi et~al.(2020)Khashabi, Min, Khot, Sabharwal, Tafjord, Clark,
  and Hajishirzi}]{khashabi2020unifiedqa}
Khashabi, D.; Min, S.; Khot, T.; Sabharwal, A.; Tafjord, O.; Clark, P.; and
  Hajishirzi, H. 2020.
\newblock UnifiedQA: Crossing Format Boundaries With a Single QA System.
\newblock arXiv:2005.00700.

\bibitem[{Kim et~al.(2018)Kim, Rohrbach, Darrell, Canny, and
  Akata}]{kim2018textual}
Kim, J.; Rohrbach, A.; Darrell, T.; Canny, J.; and Akata, Z. 2018.
\newblock Textual Explanations for Self-Driving Vehicles.
\newblock \emph{Proceedings of the European Conference on Computer Vision
  (ECCV)}.

\bibitem[{Lewis et~al.(2019)Lewis, Liu, Goyal, Ghazvininejad, Mohamed, Levy,
  Stoyanov, and Zettlemoyer}]{bart}
Lewis, M.; Liu, Y.; Goyal, N.; Ghazvininejad, M.; Mohamed, A.; Levy, O.;
  Stoyanov, V.; and Zettlemoyer, L. 2019.
\newblock BART: Denoising Sequence-to-Sequence Pre-training for Natural
  Language Generation, Translation, and Comprehension.

\bibitem[{Liu et~al.(2019)Liu, Ott, Goyal, Du, Joshi, Chen, Levy, Lewis,
  Zettlemoyer, and Stoyanov}]{roberta}
Liu, Y.; Ott, M.; Goyal, N.; Du, J.; Joshi, M.; Chen, D.; Levy, O.; Lewis, M.;
  Zettlemoyer, L.; and Stoyanov, V. 2019.
\newblock RoBERTa: A Robustly Optimized BERT Pretraining Approach.

\bibitem[{Ma et~al.(2021{\natexlab{a}})Ma, Ilievski, Francis, Bisk, Nyberg, and
  Oltramari}]{Ma2021}
Ma, K.; Ilievski, F.; Francis, J.; Bisk, Y.; Nyberg, E.; and Oltramari, A.
  2021{\natexlab{a}}.
\newblock {Knowledge-driven Data Construction for Zero-shot Evaluation in
  Commonsense Question Answering}.
\newblock In \emph{AAAI}.

\bibitem[{Ma et~al.(2021{\natexlab{b}})Ma, Ilievski, Francis, Ozaki, Nyberg,
  and Oltramari}]{ma2021exploring}
Ma, K.; Ilievski, F.; Francis, J.; Ozaki, S.; Nyberg, E.; and Oltramari, A.
  2021{\natexlab{b}}.
\newblock Exploring Strategies for Generalizable Commonsense Reasoning with
  Pre-trained Models.
\newblock \emph{EMNLP 2021}.

\bibitem[{MacQueen(1967)}]{macqueen1967classification}
MacQueen, J. 1967.
\newblock Classification and analysis of multivariate observations.
\newblock In \emph{5th Berkeley Symp. Math. Statist. Probability}, 281--297.

\bibitem[{Muppalla et~al.(2017)Muppalla, Lalithsena, Banerjee, and
  Sheth}]{muppalla2017knowledge}
Muppalla, R.; Lalithsena, S.; Banerjee, T.; and Sheth, A. 2017.
\newblock A knowledge graph framework for detecting traffic events using
  stationary cameras.
\newblock In \emph{Proceedings of the 2017 ACM on Web Science Conference},
  431--436.

\bibitem[{Radford et~al.(2019)Radford, Wu, Child, Luan, Amodei, Sutskever
  et~al.}]{radford2019language}
Radford, A.; Wu, J.; Child, R.; Luan, D.; Amodei, D.; Sutskever, I.; et~al.
  2019.
\newblock Language models are unsupervised multitask learners.
\newblock \emph{OpenAI blog}, 1(8): 9.

\bibitem[{Reimers and Gurevych(2019)}]{reimers2019sentencebert}
Reimers, N.; and Gurevych, I. 2019.
\newblock Sentence-BERT: Sentence Embeddings using Siamese BERT-Networks.
\newblock arXiv:1908.10084.

\bibitem[{Sap et~al.(2019)Sap, Le~Bras, Allaway, Bhagavatula, Lourie, Rashkin,
  Roof, Smith, and Choi}]{sap2019atomic}
Sap, M.; Le~Bras, R.; Allaway, E.; Bhagavatula, C.; Lourie, N.; Rashkin, H.;
  Roof, B.; Smith, N.~A.; and Choi, Y. 2019.
\newblock Atomic: An atlas of machine commonsense for if-then reasoning.
\newblock In \emph{Proceedings of the AAAI conference on artificial
  intelligence}, volume~33, 3027--3035.

\bibitem[{Shah et~al.(2018)Shah, Lamare, Nguyen-Anh, and
  Hauptmann}]{shah2018cadp}
Shah, A.~P.; Lamare, J.-B.; Nguyen-Anh, T.; and Hauptmann, A. 2018.
\newblock CADP: A novel dataset for CCTV traffic camera based accident
  analysis.
\newblock In \emph{2018 15th IEEE International Conference on Advanced Video
  and Signal Based Surveillance (AVSS)}, 1--9. IEEE.

\bibitem[{Song et~al.(2020)Song, Tan, Qin, Lu, and Liu}]{song2020mpnet}
Song, K.; Tan, X.; Qin, T.; Lu, J.; and Liu, T.-Y. 2020.
\newblock Mpnet: Masked and permuted pre-training for language understanding.
\newblock \emph{Advances in Neural Information Processing Systems}, 33:
  16857--16867.

\bibitem[{Wickramarachchi, Henson, and
  Sheth(2020)}]{wickramarachchi2020evaluation}
Wickramarachchi, R.; Henson, C.; and Sheth, A. 2020.
\newblock An evaluation of knowledge graph embeddings for autonomous driving
  data: Experience and practice.
\newblock \emph{arXiv preprint arXiv:2003.00344}.

\bibitem[{Williams, Nangia, and Bowman(2017)}]{williams2017broad}
Williams, A.; Nangia, N.; and Bowman, S.~R. 2017.
\newblock A broad-coverage challenge corpus for sentence understanding through
  inference.
\newblock \emph{arXiv preprint arXiv:1704.05426}.

\bibitem[{Xu et~al.(2017)Xu, Gao, Yu, and Darrell}]{xu2017end}
Xu, H.; Gao, Y.; Yu, F.; and Darrell, T. 2017.
\newblock End-to-end learning of driving models from large-scale video
  datasets.
\newblock In \emph{Proceedings of the IEEE conference on computer vision and
  pattern recognition}, 2174--2182.

\bibitem[{Xu, Huang, and Liu(2021)}]{xu2021sutd}
Xu, L.; Huang, H.; and Liu, J. 2021.
\newblock Sutd-trafficqa: A question answering benchmark and an efficient
  network for video reasoning over traffic events.
\newblock In \emph{Proceedings of the IEEE/CVF Conference on Computer Vision
  and Pattern Recognition}, 9878--9888.

\bibitem[{Zhang et~al.(2022)Zhang, Ilievski, Ma, Francis, and
  Oltramari}]{zhang2022empirical}
Zhang, J.; Ilievski, F.; Ma, K.; Francis, J.; and Oltramari, A. 2022.
\newblock An Empirical Investigation of Commonsense Self-Supervision with
  Knowledge Graphs.
\newblock \emph{arXiv preprint arXiv:2205.10661}.

\end{thebibliography}
\section{Appendix}

\subsection{Details of chosen models}
For the Inference-base reasoning model, we select models trained on the same dataset for a fair comparison, and we specifically chose MNLI because it contains 10 different genres and a very large corpus (433k).
We directly download the model from hugging-face, all of them share the same input and output format (take in a premise and hypothesis, give the confidence of the sentence pair being "entail", "neutral", "contract").
In general, we test the model's ability to make the right inference on the traffic domain without any fine-tuning. 

For KG-based reasoning models, the chosen models have two different model architectures and all have different model sizes. The encoder-only model, Roberta, uses the Masked Language Model loss to give a score for each question-candidate pair.
After getting the score for each candidate, the Roberta models are trained with the margin loss function:
$L_{margin}=\frac{1}{n} \sum_{\mbox{\tiny$\begin{array}{c}
  i=1\\
  i\neq y\end{array}$}}^{n}max(0,\eta-S_y+S_i)$,
where $S_y$ and $S_i$ are the negative averaged loss for the correct answer and the distractor respectively.
During the evaluation, the model chooses the pair with the highest score as the correct answer. The encoder-decoder model T5 learns to give an output `1' or `2' which represents the confidence of the model in predicting whether the given sentence pair is reasonable or not. For each candidate, its score is defined as $S=L_{true}-L_{false}$, then the training and evaluating process is the same as the encoder-only models.
We splice the cause and effect together as the input of the models when we test the BDD-QA-CP and BDD-QA-EP and HDT-QA datasets. 
For the HDT-QA dataset, if there is an underscore appearing in the question, we fill the underscore with candidates to make complete sentences.

For Retrieval-based models, we choose the sentence transformer version $facebook-dpr-ctx_encoder-single-nq-base$, $facebook-dpr-question-encoder-single-nq-base$ to encode the paragraphs and questions, respectively. For the unified-QA model, we use the version of Unified-QA-v2, T5-3B, which is pretrained on 20 QA datasets.

\subsection*{Implementation Details}
\label{sec:parameters}

Regarding libraries, we used python 3.7.10, PyTorch 1.9.0, and transformers 4.11.3.

Among all the supervised learning sets, we are using a learning rate of $1e^{-5}$, batch size of 32, weight decay 0.01, training epochs of 10, adam-epsilon of $1e^{-6}$, $\beta1=0.9,\beta2=0.98$, the warm-up proportion of 0.05, the margin of 1.0\\
For CPUs, we used Intel(R) Xeon(R) Gold 5217 CPU @ 3.00GHz (32 CPUs, 8 cores per socket, 263GB ram).\\
For GPUs, we used Nvidia Quadro RTX 8000.
\appendix



\end{document}